# Forgery Detection in a Questioned Hyperspectral Document Image using K-means Clustering

Maria Yaseen, Rammal Aftab Ahmed, Rimsha Mahrukh

Department of Electrical Engineering
Institute of Space Technology
1, Islamabad Highway, Islamabad, 44000, Pakistan

*Abstract—* **Hyperspectral imaging allows for analysis of images in several hundred of spectral bands depending on the spectral resolution of the imaging sensor. Hyperspectral document image is the one which has been captured by a hyperspectral camera so that the document can be observed in the different bands on the basis of their unique spectral signatures. To detect the forgery in a document various Ink mismatch detection techniques based on hyperspectral imaging have presented vast potential in differentiating visually similar inks. Inks of different materials exhibit different spectral signature even if they have the same color. Hyperspectral analysis of document images allows identification and discrimination of visually similar inks. Based on this analysis forensic experts can identify the authenticity of the document. In this paper an extensive ink mismatch detection technique is presented which uses K-Mean Clustering to identify different inks on the basis of their unique spectral response and separates them into different clusters.**

*Keywords— Hyperspectral Imaging; Ink Mismatch Detection; K-Means Clustering.*

## I. Introduction

Every object is capable of transmission, reflection and absorption of light in a specific manner while these properties gives it a specific color. Basically, this is referred to as spectral response. Human eye can detect various differences in different colors but sometimes it is hard to recognize between two colors which are actually different but we are unable to recognize it.

Topic discussion of our paper is Hyperspectral imaging (HSI), which is a technique that analyzes a wide spectrum of light instead of just assigning primary colors (red, green, blue) to each pixel. [1] The light striking each pixel is broken down into many different spectral bands in order so it can give more information on what is imaged. The algorithms and the image processing methodologies associated with HSI are a product of military research, and were primarily used to identify targets and other objects against background clutter. In the past, HSI has also seen applications like civil and has also been useful in satellite technology. For future it might become an inexpensive and quick tool for the assessment of tissue conditions at the time of diagnosis and during surgery.

Hyperspectral document imaging could be done by a non-destructive apparatus. It permits recording and analyzing of the documents in hundreds of barely dispersed spectral groups, hence uncovering the hidden details within the scene of interest without getting in direct contact with it. HSI bears a huge potential for accurate separation of materials based on their unique spectral marks. R. J. Hejdam et al. [2] used HSI analysis for restoration of old reports,
and focused on the practical scenarios where ink might not be identified by naked eyes. F. Hollaus et al. [3] proposed picture improvement for debased pictures by
utilizing the spectral as well as spatial data. A. Abbas et al [4] proposed a novel unsupervised clustering method. Another method [5] introduced is CNN based novel method for ink mismatch detection in hyperspectral images which detects ink mismatch by classifying the ink pixels based on their spectral responses.

In this paper, an ink examination procedure based on hyperspectral is proposed for no. of inks used in a folder detection. Our primary Centre is to recognize visually similar inks which are blended in changing extents to make an unbalance clustering issue. We have performed a task where we had to display different bands and then also recognize the no. of inks that has been used in that particular HSI. We first developed a program on python and showed 1st, 10th, 30th bands. We performed K-means clustering which helped us find out no. of inks that were used in that document. K-means clustering is one of the simplest and popular unsupervised machine learning algorithms. Typically, unsupervised algorithms make inferences from datasets using only input vectors without referring to known, or labelled, outcomes.

This resulted in plotting of different colors in our graph which identified no. of inks that were used in our document. Further we separated our background to plot foreground, since the writing written on document was considered as foreground so it was easy to identify no. of inks. K-means algorithm helped us display our image in segmented form which resulted in different colors in the writing. Also, different colors were observed in the spectral response which helped us in detecting no. of inks in our document.

## II. RELATED WORK

Hyperspectral image analysis has considerably developed the efficiency of forgery detection systems over the recent years. Various HSI based techniques for automated forgery detection are proposed in literature. E.B. Brauns et al. [6] developed a hyperspectral sensing system for non- destructive forgery detection in questioned documents by employing an interferometer. A comparatively complex and advanced hyperspectral imaging framework for historic document examination was developed at the National Archives of Netherlands that provided high resolution in the spatial as well as the spectral domain, i.e. ranging from near ultraviolet to near infrared [7]

Hyperspectral imaging has developed as an effective nondestructive tool for improving readability of extremely faded documents [8], ink aging and in forensic document analysis [9]. Hedjam et al. [10] proposed algorithm capable of improving the visual quality of degraded image based on multispectral imaging. Hedjam et al. [11] also proposed a mathematical model for improving the readability of extremely deteriorated text. Hollaus et al. [12] introduced a method for enhancing degraded and ancient writings captured by multispectral imaging system on the basis of spectral as well as spatial information. Brauns et al. [6] designed a hyperspectral imaging technique based on Fourier transform for the non-destructive analysis of potentially duplicitous documents. A more sophisticated hyperspectral imaging system for quantifying and monitoring aging processes of documents was developed by National Archives of Netherlands [13]. Morales et al. [14] proposed an approach for ink analysis in pen verification and handwritten documents using Least Square SVM classification. Silva et al. [15] developed a non-destructive method to detect fraud in documents based on different chemo metric techniques. Z. Khan et al. [16] proposed a joint sparse band selection based hyperspectral imaging document analysis technique to distinguish different metameric inks. A. Abbas [17] proposed an ink analysis technique based on hyper- spectral un-mixing ink mismatch detection, their main focus was to distinguish visually similar inks which are mixed in varying proportions to form an unbalance clustering problem. M.J Khan [18] proposed an efficient automatic ink mismatch detection technique using multispectral image analysis. Ink pixels were segmented out using local thresh-holding and Fuzzy C-Means Clustering (FCM) was used to divide the spectral response vectors of ink pixels into different clusters which relates to different inks used in the document. R. Qureshi [19] explored the potential of HSI for document image analysis and presented a comprehensive review of the literature and future prospects. The challenges involved in the acquisition and processing of hyperspectral document images were also presented.

## III. METHODOLOGY

We were provided a hyperspectral document image with 33 bands for detection of the number of inks being used in it. First of all we downloaded the zip file containing all the bands and saved them into a folder.
The paths of the required bands were used to store them in a variable which were then converted to grayscale before display. The three bands ($1^{st}$, $10^{th}$ and $30^{th}$) in the grayscale format were then plotted. The following is the output of this task which was performed solely to check the correct display of bands of the document image.

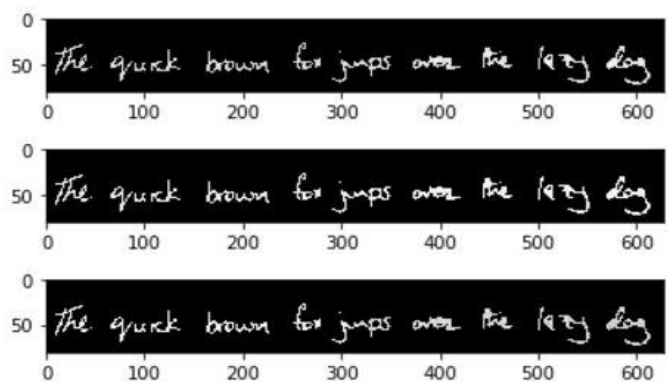

*Figure 1 Bands no. 1, 10 and 30 of the hyperspectral document image*

The next major step was that the spectral response of the foreground pixels were to be plotted. As already discussed above that each object has different spectral behavior so plotting the spectral responses of the ink pixels would tell us the number of distinct inks being used in the document.

For this task, we saved all the 33 bands in a single variable by giving the path of each band one by one and then reading and storing them in an array which had all the 33 bands. This new variable now had a complete hyperspectral image. This was the first step performed before plotting the spectral responses. Since, we were to plot the response of the foreground pixels, which means we had to ignore the background. As we all know that background pixel has a value 0 so we first read the complete image and then plotted all the pixel values greater than zero, in other words the foreground pixels were plotted. The plot showed the response of the foreground of the document. The output of this task is shown below.

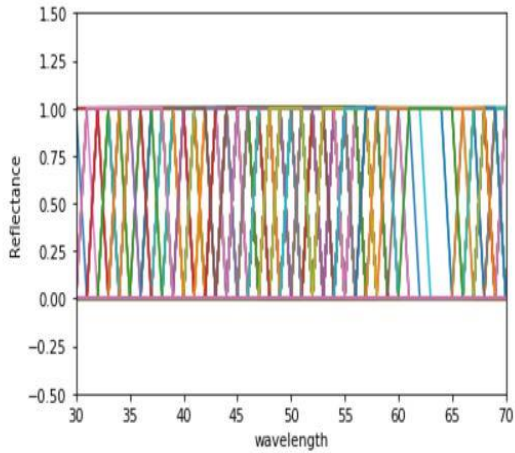

*Figure 2 Spectral response of foreground pixels*

The limits of x and y axis were adjusted accordingly to make the graph visually understandable. By looking at the foreground response we concluded that there are approximately 5 distinct colors in the plot. These distinct colors corresponded to the distinct inks used to write the document image. This part of the task laid the foundation of the next task because we were able to figure out the number of clusters which were to be made.

For next step we were to find the number of inks used in the document. Since we had already concluded the number, the only task we had was to use an efficient pattern recognition tool to classify the document and verify the number of inks used. We had a vast choice to use clustering techniques such as C-means, fuzzy C-means, SVM, PCA etc. But we opted for K-means clustering because it is one of the simplest technique yet gives efficient results. In K-means clustering we define the number of clusters we have to divide our image into initially and then the program computes the random centroids for those clusters and assign the pixels according to their distance from those centroids and keep updating the centroids and clusters until there is no further change. Another reason for opting K-means clustering was that it is a well-known technique so any irregular or unexpected outcome would have been easily detected by us.

As far as implementation is concerned, before applying K-means clustering on our document image, we had to do thresholding to separate the background from the foreground.

For thresholding, we did local thresholding; binary thresholding which converts the image into two distinct colors; black and white image representing the background and foreground respectively.

We began with reading the document image, and then applied binary threshold on to it with a threshold value of 40 which means all the value of pixels below this threshold were made part of the background. The choice of this threshold was done by hit and trial, we checked for all values until a proper binary output was received. We then displayed the original and the binary threshold image.

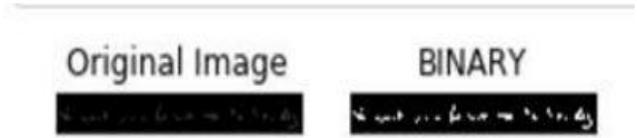

*Figure 3 Original and Binary threshold Image*

Once, the thresholding was done, we applied the K-means clustering onto the image with the number of clusters equal to five (5). The k-means clustering was completed after the iterations ended. To verify the correctness of this clustering, we plotted the segmented image which is shown below. The segmented image had a plot consisting of five different colors representing the number of clusters in which the document image has been clustered. This basically, verified that the number of inks used in the document were five.

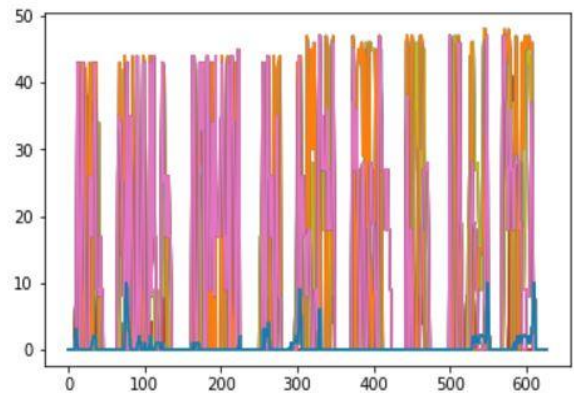

*Figure 4 Plot of the Segmented Image*

In the last portion we had to show the document written with different inks.

For this, we simply displayed the segmented image which was achieved in the previous part. The following figure shows the color labelled document image.

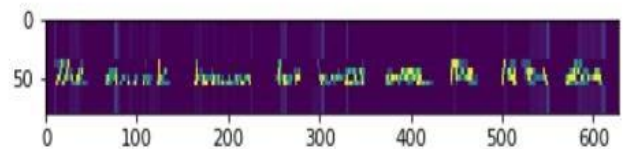

*Figure 5 Final Resultant Image*

In this image the different inks were displayed with different colors so that it can be easily distinguished that what portion is written with a distinct ink.

## IV. EXPERIMENTAL ANALYSIS AND RESULTS

Number of inks present in hyperspectral document image are identified through unique spectral response of ink pixels. Furthermore, image was segmented out to background and foreground, since the document image used for analysis is considered as foreground. K-means clustering algorithm helped us display our image in segmented form which resulted in different colors in the writing.

The method used for clustering is considered to be one of the simplest clustering methods and was used for ease yet it gave efficient results. However, a complex and better algorithm would have also resulted in the same outputs may be with better precision. Since, the working of K-means was well known to us, as beginners in aspect of hyperspectral image processing we preferred using it. We see the majority of ink pixels are correctly clustered.

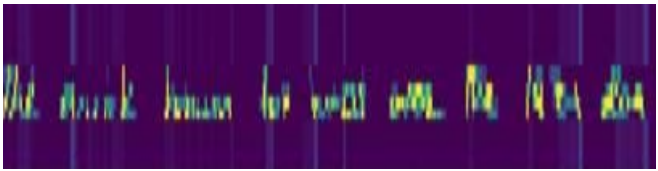

*Figure 6 A close-up of the correctly clustered document image*

The proposed method provides better discrimination between inks in a questioned document. The experiment was performed on a machine with Intel Core i5 @ 2.40GHz with 6.00GB RAM using Python3.8.

## V. CONCLUSION

Hyperspectral imaging helps to view the image as different bands and the detection of forgery becomes easier by the use of this non-destructive ink mismatch detection technique. The proposed method uses a simple pattern recognition tool to extract the number of different inks used in the document. We expect that these findings can further be improved and polished. The use of clustering algorithms is thus very beneficial in the domain of forgery detection in questioned documents.

## ACKNOWLEDGMENT

We are thankful to Sir Khurram Khurshid for his help and encouragement during the project.